\documentclass[acmsmall]{acmart}

\AtBeginDocument{%
  \providecommand\BibTeX{{%
    \normalfont B\kern-0.5em{\scshape i\kern-0.25em b}\kern-0.8em\TeX}}}

\acmConference[ETRA'24]{ETRA 2024}{June 04--07,
  2024}{Glasgow, United Kingdom}

\begin{document}

\title [SAM for Eye Image Segmentation] {Zero-Shot Segmentation of Eye Features Using the Segment Anything Model (SAM)}

\author{Virmarie Maquiling}
\email{virmarie.maquiling@tum.de}
\affiliation{%
	\institution{Human-Centered Technologies for Learning, Technical University of Munich}
	\streetaddress{Marsstraße 20-22}
	\city{Munich} 
	\country{Germany} 
	\postcode{80335}
}

\author{Sean Anthony Byrne}
\email{sean.byrne@imtlucca.it}
\affiliation{%
	\institution{MoMiLab, IMT School for Advanced Studies Lucca}
	\streetaddress{Piazza S.Francesco, 19}
	\city{Lucca, LU} 
	\country{Italy} 
	\postcode{55100}
}

\author{Diederick C. Niehorster}
\affiliation{%
  \institution{Lund University Humanities Lab \& Dept. of Psychology, Lund University}
  \city{Lund}
  \country{Sweden}}
\email{diederick_c.niehorster@humlab.lu.se}

\author{Marcus Nystr{\"o}m}
\affiliation{%
  \institution{Lund University Humanities Lab, Lund University}
  \city{Lund}
  \country{Sweden}}
\email{marcus.nystrom@humlab.lu.se}

\author{Enkelejda Kasneci}
\email{enkelejda.kasneci@tum.de}
\affiliation{%
	\institution{Human-Centered Technologies for Learning, Technical University of Munich}
	\streetaddress{Marsstraße 20-22}
	\city{Munich} 
	\country{Germany} 
	\postcode{80335}
}

\renewcommand{\shortauthors}{Maquiling and Byrne, et al.}

\begin{abstract}
The advent of foundation models signals a new era in artificial intelligence. The Segment Anything Model (SAM) is the first foundation model for image segmentation. In this study, we evaluate SAM's ability to segment features from eye images recorded in virtual reality setups. The increasing requirement for annotated eye-image datasets presents a significant opportunity for SAM to redefine the landscape of data annotation in gaze estimation. Our investigation centers on SAM's zero-shot learning abilities and the effectiveness of prompts like bounding boxes or point clicks. Our results are consistent with studies in other domains, demonstrating that SAM's segmentation effectiveness can be on-par with specialized models depending on the feature, with prompts improving its performance, evidenced by an IoU of 93.34\% for pupil segmentation in one dataset. Foundation models like SAM could revolutionize gaze estimation by enabling quick and easy image segmentation, reducing reliance on specialized models and extensive manual annotation.
\end{abstract}


\setcopyright{acmlicensed}
\acmJournal{PACMCGIT}
\acmYear{2024} \acmVolume{7} \acmNumber{2} \acmArticle{1} \acmMonth{6}\acmDOI{10.1145/3654704}

\ccsdesc[100]{Human-centered computing~Virtual reality}
\ccsdesc[500]{Computing methodologies~Image segmentation}
\ccsdesc[300]{Computing methodologies~Neural networks}


\keywords{Eye-tracking, Segmentation, Segment Anything Model, Zero-shot learning, Foundational models, Prompt Engineering}

\begin{teaserfigure}
    \centering
    \includegraphics[width=0.7\textwidth]{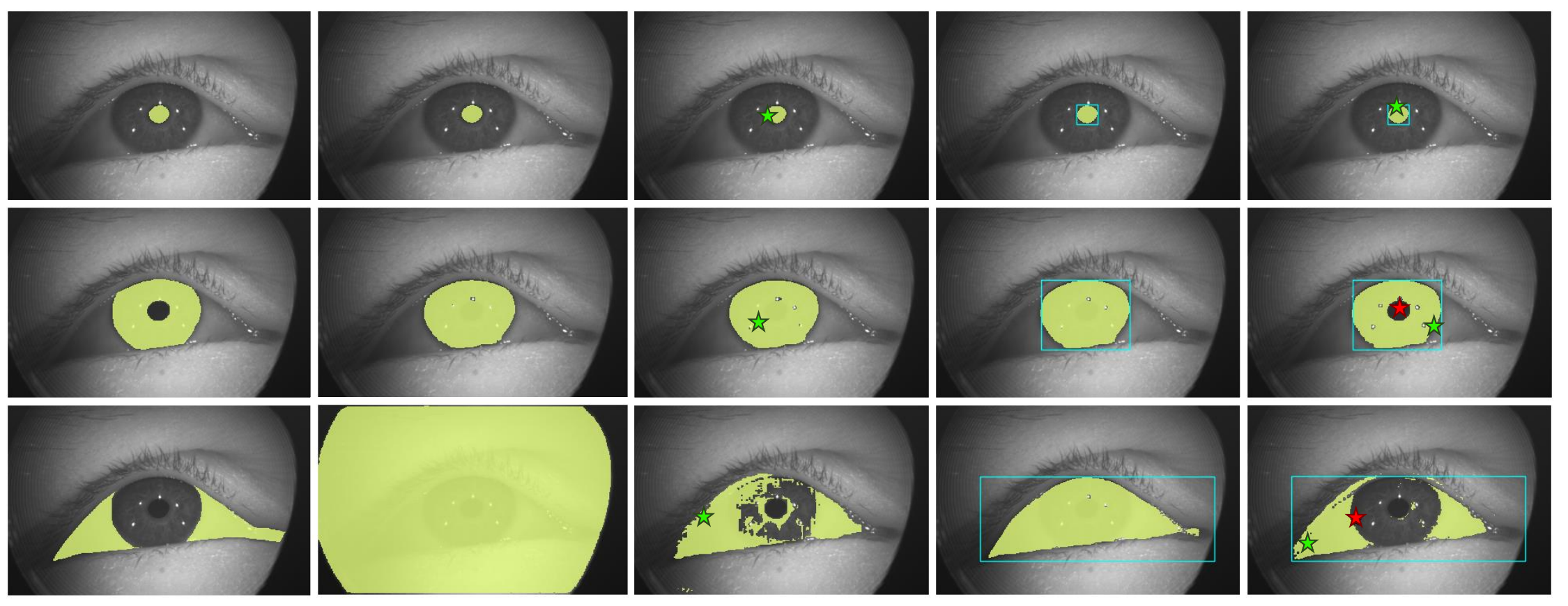}
    \caption{Visualization of SAM's segmentation performance according to strategy. On the leftmost column are the ground truth masks for key eye features: pupil, iris, and sclera. The remaining columns show representative prompt strategies used in the experiment overlayed with the resulting masks generated by SAM. Green and red stars represent foreground and background point prompts. The light blue rectangle represents a bounding box prompt. Image source from OpenEDS2020 dataset \cite{palmero2021openeds2020}.
    }
    \label{fig:qualitative}
    \Description{Visualization of SAM's segmentation performance}
\end {teaserfigure}

\maketitle

\section{Introduction}
Foundation models and Generative AI are already being celebrated as the groundbreaking developments of the decade in the realm of artificial intelligence. They mark a significant paradigm change in the way users interact, develop, and deploy deep learning models~\cite{bommasani2022opportunities}. Compared to traditional deep learning models, foundation models have scalable attributes, such as the vast number of trainable parameters and the extensive amounts of data they are trained on. Due to these scaling attributes foundational models demonstrate a remarkable ability to adjust to downstream tasks and exhibit proficiency on data distributions previously unseen during their training phase~\cite{kirillov2023segment,bommasani2022opportunities}. From a non-technical standpoint, these models have made it easier to utilize tools powered by artificial intelligence, managing complex tasks that previously required specialized supervised learning models, and often custom datasets collected specifically for the problem at hand~\cite{bommasani2022opportunities,zhou2023comprehensive}. In contrast, foundation models simplify the deployment process through their innate zero-shot learning capabilities—where the model makes accurate predictions on tasks without having seen any examples during training~\cite{pmlr-v37-romera-paredes15}. This potential can be further improved with straightforward prompts. These prompts can be as easy to create as a single mouse click or circling a significant feature within an image, thereby diminishing the complexity and time needed to apply deep learning models on a large scale~\cite{mazurowski2023segment}.

In this study, we harness the power of the Segment Anything Model (SAM)~\cite{kirillov2023segment}, the first promptable foundation model developed for image segmentation, and which has demonstrated remarkable abilities within a variety of different domains, including both natural and specialized forms of imagery, such as medical imaging~\cite{ma2024segment}, and remote sensing~\cite{archit2023segment}. We apply SAM in the context of gaze estimation, where we task the model to segment eye regions in near-eye images obtained from a virtual reality (VR) setting. We investigate various prompting strategies to bolster SAM's performance and closely examine its effectiveness in identifying three specific ocular regions: the pupil, iris, and sclera. Annotating eye regions, a well-known task within video-based eye tracking, is both subjective and costly in terms of time. These segmented features are essential not only for precise gaze estimation but also as training data for specialized deep learning models that require numerous annotated examples in order to make accurate predictions.

In addition to their ease of deployment, foundation models like SAM promise to overcome the domain generalization issue—a notable obstacle across machine learning, where models typically falter when applied to data different from their training set~\cite{zhou2022domain}. This challenge is particularly evident in gaze estimation, where it has been noted that many supervised learning models, such as convolutional neural networks, often fail to generalize their understanding of eye components, resulting in subpar performance on data not encountered during their training due to differences in recording setup such as lighting, camera angle, and any physiological differences across participants~\cite{kothari2022ellseg,byrne2023leyes, kim2019nvgaze}. The development of extensive datasets recorded across multiple devices~\cite{Fuhl_2021, kothari2022ellseg} and the generation of synthetic data~\cite{kim2019nvgaze,nair2020rit,maquiling2023v, byrne2023leyes} have been suggested to bypass the domain generalization problem in gaze estimation. Yet, these approaches have not completely resolved the issue and also introduce significant entry barriers, including the need for substantial computational power and specialized expertise in computer vision. However, foundation models have showcased exceptional generalization abilities in zero-shot and few-shot scenarios, adeptly adapting to unseen tasks and data distributions~\cite{brown2020language,bommasani2022opportunities}.

Our investigation is threefold. First, we provide a literature review on foundation models, zero-shot learning, and eye image segmentation to give the reader a clear understanding of the problem and potential of foundation models in this domain. Second, we explain what it means for a model like SAM to be ``promptable'' and introduce a set of prompting strategies designed to explore SAM's capabilities for eye image segmentation. Third, using these tailor-made prompting strategies, we evaluate SAM's performance on two public eye image segmentation datasets, providing quantitative and qualitative analysis with the aim of illuminating the potential and limitations of applying SAM to the domain of eye tracking. Our results reveal that SAM, in a zero-shot learning context, can produce annotations that rival manual annotations derived from experts and annotations from traditional supervised learning models in segmenting the pupil, even with minimal manual guidance. Nevertheless, the model exhibits certain limitations in the accurate annotation of the sclera and, to a lesser degree, the iris. We have made all code available to users at the following repository for both benchmarking and labeling their own datasets: https://github.com/vbmaq/ET-SAM. 

\section{RELATED WORK}

\subsection{An Introduction to Foundation models}
Foundation models are a relatively new concept in deep learning, and as such, there is not an extensive amount of literature available on them yet. Although these models first gained prominence in the realm of Natural Language Processing (NLP), they have recently made significant strides in the field of computer vision. Foundation models are large-scale machine-learning models trained on massive amounts of data that can be adapted for a wide range of downstream tasks. These models mark a departure from conventional deep learning approaches due to their scale, boasting not only an unprecedented number of parameters (up to hundreds of billions) but also being pre-trained on extensive datasets. This enables them to adapt to various tasks through fine-tuning and exhibit emergent behaviors, such as strong performance on previously unseen data distributions~\cite{bommasani2022opportunities}. Leading the way are models like BERT~\cite{devlin2018bert}, Generative Pre-Trained Transformers (GPTs)~\cite{brown2020language}, and CLIP~\cite{radford2021learning}, all of which leverage large-scale data for training. ChatGPT, fine-tuned from the GPT-3 model with its 175 billion parameters, excels in various tasks through natural language prompts and has significantly influenced  AI over the last year~\cite{zhou2023comprehensive}. Already, there is a growing trend toward the creation of specialized foundation models, such as MedSAM for medical imaging~\cite{ma2024segment}, and even more domain-specific,  RETFound, a model designed specifically for the analysis of retinal images~\cite{zhou2023foundation}. For a comprehensive review of foundation models, see ~\cite{bommasani2022opportunities}.

\subsection{The Segment Anything Model } 
\label{sammeth}
The Segment Anything model (SAM)~\cite{kirillov2023segment} is a state-of-the-art vision transformer trained on the SA-1B dataset, a collection of diverse images (11 million) accompanied by high-quality segmentation masks (1.1 billion) created specifically for SAM's training, making it the largest publicly available image segmentation dataset to date. The dataset was generated in three stages. In the first stage, a set of images was labeled by a team of professional annotators by clicking on foreground and background points and refining masks generated by an early version of SAM that was trained on common public segmentation datasets. In the second stage, SAM automatically generated masks of a subset of the dataset, while the annotators focused on annotating the rest of the objects to increase mask diversity. In the last stage, the remaining masks were automatically generated by SAM by prompting the model with a $32\times32$ grid of foreground points~\cite{kirillov2023segment}. 

To enable zero-shot generalization in image segmentation tasks, a novel promptable segmentation task was introduced ~\cite{kirillov2023segment}, in which, given an image and any prompt, the goal is to return a valid segmentation mask. In this context, a ``prompt'' specifies which object needs to be segmented by the model, e.g., by providing spatial information. This can be manually given by a user or generated by a different model (e.g., using an object detection model). 
To support this novel promptable task, the model, as depicted in Fig. \ref{fig:sammodel}, consists of three main components (for a more detailed description, refer to the original paper \cite{kirillov2023segment}): 

\begin{figure}[h]
    \centering
        \includegraphics[width=1\linewidth]{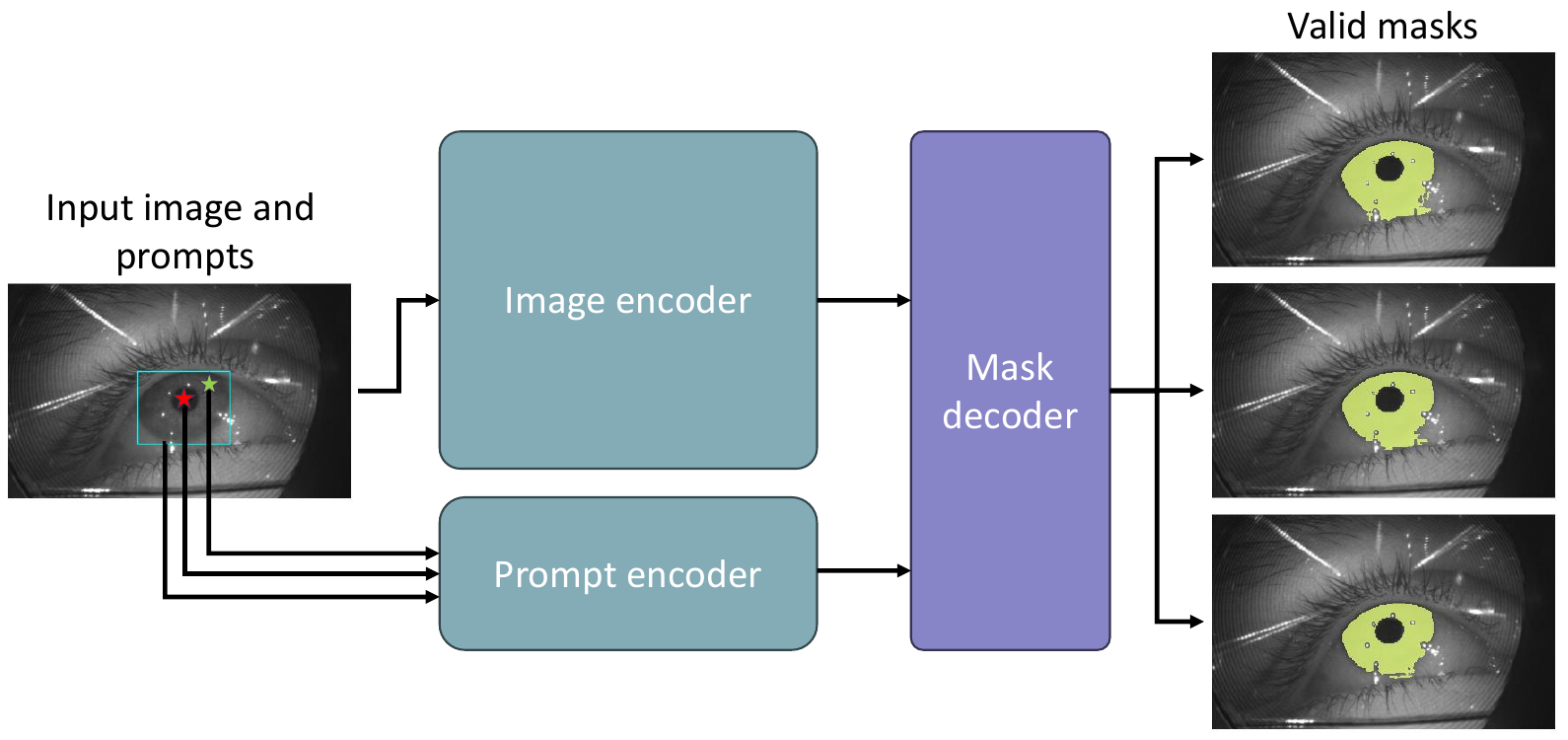} 
    \caption{A high-level schematic of the Segment Anything Model (SAM) featuring an image encoder, a prompt encoder, and a lightweight mask decoder. The image is fed to the image encoder, while a set of prompts (in this example, depicted as point- and bounding box prompts drawn on the image) are fed to the prompt encoder. The embeddings generated by the image and prompt encoders are passed to the mask decoder which outputs valid masks with varying confidence scores. Alternatively, it may output a single mask averaged over all valid masks. Image source from \cite{palmero2021openeds2020}.}
    \label{fig:sammodel}
\end{figure}

\begin{itemize}
    \item \textbf{Image encoder}: This uses a Vision Transformer \cite{dosovitskiy2010image}, pre-trained using Masked Autoencoders (MAE)~\cite{he2022masked}, as its backbone and can be run once per image before prompting the model. The image encoder uses an input resolution of $1024\times1024$ by resizing and padding the input image. It then outputs a $16\times$ downscaled image embedding of size $64\times64$. 
    \item \textbf{Prompt encoder}: This encoder accepts sparse (points, boxes, text) and dense (masks) prompts and encodes them as positional embeddings (points and boxes), by CLIP~\cite{radford2021learning} (text), or through convolutions (masks). Point prompts are given as pixel coordinates in the image, accompanied by point labels indicating whether each corresponding point is in the foreground or the background. Bounding box prompts consist of a pair of pixel coordinates indicating the top-left corner and the bottom-right corner of the box. For information regarding other prompt modes, refer to the original paper \cite{kirillov2023segment}.
    \item \textbf{Mask decoder}: This lightweight decoder maps the image and prompt embeddings to an output mask. 
\end{itemize}

SAM supports two modes: an automatic ``everything'' mode, in which the user simply inputs an image to SAM, which then returns all predicted masks found within the image, and a manual mode which expects the user to input an image and a combination of manual prompts to indicate the location of the object to be segmented, e.g. by simply clicking on the image or drawing a bounding box around the region of interest. 
Three pre-trained models of varying network sizes are available in SAM's GitHub repository\footnote{https://github.com/facebookresearch/segment-anything/}. From smallest to largest, they are named ViT Base (ViT-B), Vit Large (ViT-L), and Vit Huge (ViT-H). The authors report substantial improvement of ViT-H over ViT-B, while showing only marginal improvement over ViT-L. However, due to its increased complexity, inference time is also multiplied \cite{kirillov2023segment}. Nevertheless, previous work reported that ViT-H did not show a significant advantage over ViT-B, with ViT-B even surpassing larger variants in some segmentation tasks \cite{huang2023segment, mattjie2023zeroshot}.

\subsection{Applications of the Segment Anything Model}
 The application of the Segment Anything Model (SAM) has already been broad, extending to various domains and specialized tasks, with adaptations that have seen it being used in video tracking~\cite{yang2023track} and three-dimensional environments~\cite{shen2023anything}. However, this literature review section will focus exclusively on the downstream performance of the original SAM architecture. To our knowledge, only a single study exists where SAM has been integrated with eye-tracking data: GazeSAM~\cite{wang2023gazesam}. This model enhances the automation of medical image segmentation by capturing the eye movements of radiologists and using this gaze data to prompt SAM for interactive segmentation tasks. SAM has been notably utilized in medical imaging in many studies. One study demonstrates SAM's versatility across 19 different medical imaging datasets by employing interactive point and box prompts in a zero-shot segmentation task. The outcomes showed variable Intersection Over Union (IOU) scores depending on the specific dataset and segmentation task, with scores fluctuating from 0.1135 in spine datasets to 0.8650 in MRI images~\cite{mazurowski2023segment}. These findings were largely corroborated in a separate study, emphasizing SAM's capacity to minimize annotation time in medical image analysis~\cite{huang2023segment}. This study also concluded that SAM benefits from point and box prompts to enhance object detection in medical images, yet much like in the aforementioned work~\cite{mazurowski2023segment} its performance remains inconsistent across different objects and modalities. %
 Notably, in a related study, box prompts were identified as more robust than point prompts, consistently outperforming them even when subjected to added perturbance \cite{mattjie2023zeroshot}.
 Taking this a step further,  researchers created MedSAM, a foundational model specifically for medical images~\cite{ma2024segment}. The study leveraged the SAM dataset to pre-train a transformer model, which was then fine-tuned using over 1 million medical images. The MedSAM model not only surpassed the standard SAM in performance but also matched or exceeded specialized models, like a U-net, which are tailored for particular medical imaging tasks~\cite{ma2024segment}. Outside of medical imaging, SAM has been applied in domains such as graphs~\cite{jing2023segment}, remote sensing~\cite{wang2023scaling} and microscopy~\cite{archit2023segment} all with various degrees of success depending on the task. 

\subsection{Segmentation of eye-images recorded using video-based eye-tracking systems}

In the context of image analysis for video-based eye-tracking, segmentation is an essential process that typically involves the detection and segmentation of areas such as the pupil, sclera, iris, and any corneal reflections according to the specific requirements of the task. Pupil center localization is a core part of many gaze estimation methodologies~\cite{kim2019nvgaze}. Additionally, techniques that track both the pupil and corneal reflections (P-CR eye tracking) are widely employed for gaze estimation, representing the leading approach in the video-based eye-tracking sector~\cite{dunn2023minimal}. Limbus tracking, which uses the boundary between the iris and sclera, is another technique utilized for gaze and eye movement monitoring~\cite{holmqvist2011eye}.

Historically, the segmentation of eye images from video footage was often done using computer vision techniques, such as ellipse fitting or thresholding methods. These techniques rely on handcrafted algorithms and heuristics to identify features~\cite{kothari2022ellseg, santini2018pure}. Despite their effectiveness, they can fail in the presence of, for instance, occlusions, changing illumination conditions, or reflections in the eye images~\cite{fuhl2016pupil}. A potential remedy has been the adoption of machine learning and, subsequently, deep learning models, which have considerably enhanced the robustness of gaze estimation algorithms~\cite{kothari2022ellseg, fuhl2016pupilnet, fuhl2020tiny}. The issue of data scarcity and the aspiration to tackle domain generalization issues have led to propositions for both large-scale~\cite{Fuhl_2021, kothari2022ellseg} and synthetic datasets~\cite{byrne2023precise,byrne2023leyes,kim2019nvgaze,maquiling2023v,nair2020rit}. These are intended as solutions, yet the challenge of domain generalization remains largely unaddressed in the literature. To date, the necessity for a new specialized model, which requires significant expertise to construct, persists without a convincing alternative.

\section{Methodology}

\subsection{Gaze Datasets}

\begin{figure}[h]
    \centering
        \includegraphics[width=0.5\linewidth]{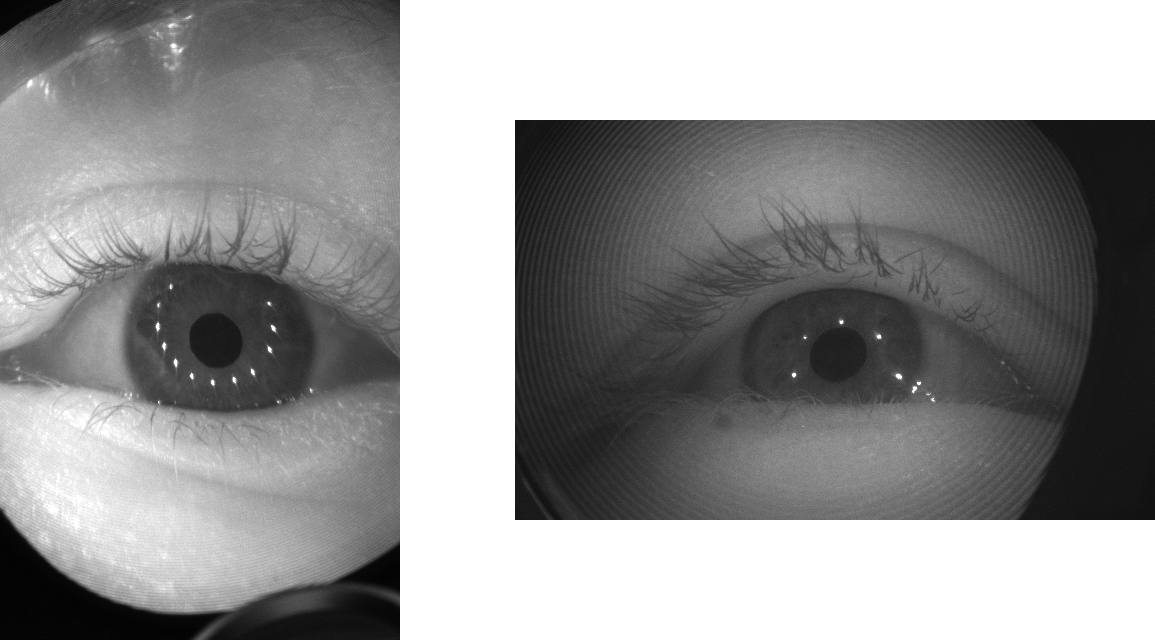} 
    \caption{Representative images from OpenEDS2019 \cite{garbin2020dataset} (left) and OpenEDS2020 \cite{palmero2021openeds2020} (right) datasets. Both datasets are captured using a VR head-mounted display equipped with dual synchronous eye-facing cameras. The OpenEDS2019 dataset contains images with a resolution of $400\times640$ while the images from the OpenEDS2020 dataset have a resolution $640\times400$.}
    \label{fig:side_by_side}
\end{figure}

To assess SAM's efficacy, we utilized two datasets: the Open Eye Challenge dataset (OpenEDS2019)~\cite{garbin2020dataset}, which is widely used to benchmark gaze estimation methods~\cite{nair2020rit, chaudhary2022temporal, kim2019nvgaze, kothari2021ellseg, kothari2022ellseg, byrne2023leyes}, and its successor, the OpenEDS 2020 Challenge dataset (OpenEDS2020)~\cite{palmero2021openeds2020}. OpenEDS2019~\cite{garbin2020dataset} was collected using a VR head-mounted display (HMD) equipped with two synchronized eye-facing cameras under controlled illumination at a frame rate of 200 Hz. The dataset encompasses eye-region video footage from 152 individual participants, comprising a total of 12,759 images at a resolution of $400\times640$ pixels, coupled with pixel-level annotation masks for key eye-regions, namely the pupil, iris, and sclera. For a complete data description, refer to the original paper~\cite{garbin2020dataset}. The OpenEDS 2020 Challenge dataset~\cite{palmero2021openeds2020} consists of eye-image sequences recorded through a VR HMD mounted with dual eye-facing cameras under controlled lighting conditions at a frame rate of 100 Hz and recorded from 80 participants while performing various gaze-related tasks. It is further separated into two subsets: 1) Gaze Prediction Dataset and 2) Eye Segmentation Dataset. The latter consists of 200 sequences sampled at 5 Hz accompanied by manually annotated segmentation labels for 5\% of the dataset, totaling 2605 images at a resolution of $640\times400$ pixels. A representative image from each dataset is displayed in Fig. \ref{fig:side_by_side}. Before being fed into SAM's image encoder, each image is converted to RGB with a pixel intensity range of $[0, 255]$. No further adjustments to the images were performed. Both datasets contain partially closed eyelids, which are not excluded in our evaluation; instead, we established exclusion criteria tailored to specific features, as discussed in Section \ref{sec:manualprompting}.

As part of each challenge, the challenge authors conducted baseline experiments to evaluate performance on both datasets. Specifically, they trained an encoder-decoder neural network derived from the SegNet architecture~\cite{badrinarayanan2017segnet} for each dataset. For OpenEDS2019, the model was trained for 200 epochs on the training set provided in the challenge dataset, achieving a mean intersection over union (mIoU) score of 91.4\% on the test set consisting of all three features. For OpenEDS2020, the authors trained the model for 150 epochs on the subset of images with corresponding segmentation masks, reporting a mIoU score of 84.1\% 

\subsection{Prompting Strategies}

\begin{figure}
    \centering
    \includegraphics[width=\linewidth]{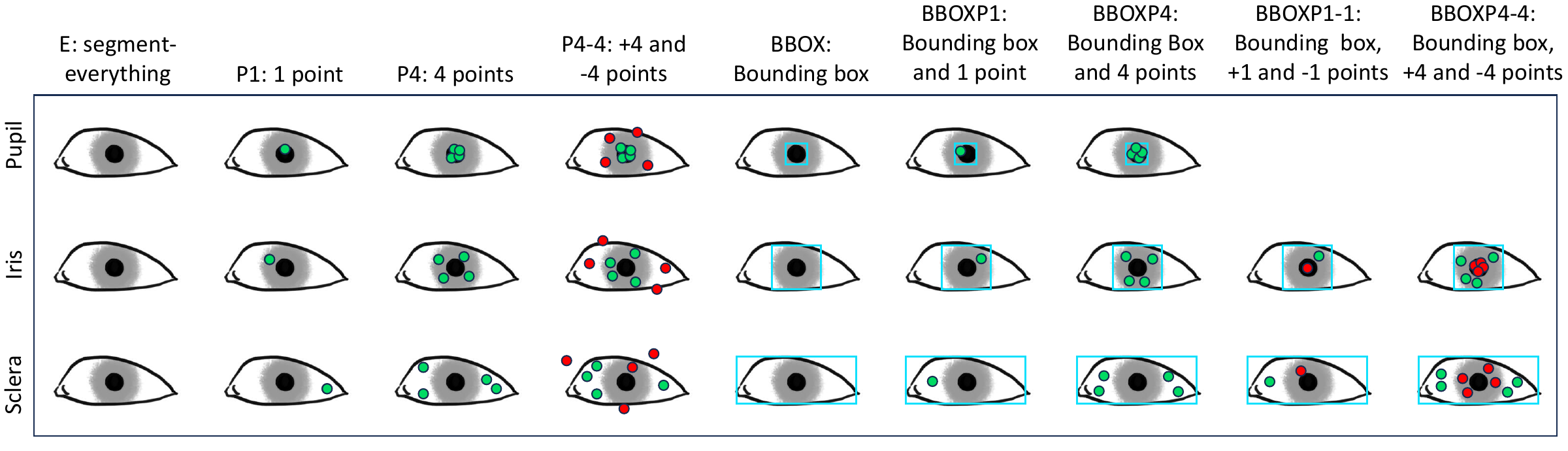}
    \caption{Prompt strategies for segmenting pupil, iris, and sclera. Green points represent foreground point prompts, while red points represent background point prompts. Bounding box prompts are visualized as light blue rectangles surrounding the feature of interest.}
    \label{fig:prompts}
\end{figure}

We assess SAM's performance on eye images, utilizing both the automatic ``everything'' mode and the manual prompting mode and employing a combination of bounding box and point prompts. SAM's main strength lies in its capability to incorporate prompts, making it imperative to evaluate the impact of different prompt strategies on model performance. We designed a set of prompting strategies that could be employed by a human annotator using SAM for eye image annotation, as depicted in Fig. \ref{fig:prompts}. Each prompt strategy is described below.

\subsubsection{Automated/no prompting} 
The SAM model has the option to automatically segment the entire input image into distinct segmentation regions. SAM outputs a list over results represented by dictionaries containing the predicted binary masks, the predicted quality of each mask, and the low resolution logits for each mask. In the context of data annotation, upon feeding the image to the model, the user receives a series of masks as output. However, not all these segmentations may align with the user's needs. Consequently, the user may opt to manually choose the masks that best align with the features relevant to them. To simulate this, we adopt the mask-matching mechanism proposed by \citeauthor{huang2023segment}~[\citeyear{huang2023segment}], which involves identifying masks with the closest Dice scores to the ground truth masks representing the pupil, iris, and sclera. Subsequently, the masks with the highest scores are considered as the model's predicted features for the given image. We refer to this strategy as E (everything) in our results. 

\subsubsection{Manual prompting}
\label{sec:manualprompting}
Alternatively, SAM can be prompted to segment only a specific portion of an image. To achieve this, SAM gives users the flexibility to use diverse prompt strategies to select the feature to segment. Possible strategies include clicking within or outside the region of interest, drawing a loose or tight bounding box around it, or a combination of these approaches. While an exhaustive comparison of all potential prompt strategies is unfeasible, we can categorize these strategies into the following actions: using a point prompt, using multiple point prompts, using a bounding box, and a combination thereof. In this mode, we specifically focus our analysis of SAM's performance on the following prompt designs, visually depicted in Fig. \ref{fig:prompts}: 
\begin{enumerate}
    \item \textbf{Point prompts (P1, P4)}: single or multiple (4) points are randomly positioned anywhere on the ground truth mask.
    \item \textbf{Positive and negative point prompts (P4-4)}: taking inspiration from \citeauthor{huang2023segment}~[\citeyear{huang2023segment}], 4 positive (foreground) points are randomly positioned on the ground truth mask. A bounding box around the ground truth mask is then calculated and doubled in size. 4 negative points are placed outside the ground truth mask within the resulting bounding box. 
    \item \textbf{Bounding Box (BBOX)}: a tight bounding box surrounding the ground truth mask.
    \item \textbf{Bounding Box with point prompts (BBOXP1, BBOXP4)}: a tight bounding box surrounding the ground truth mask and a single or multiple (4) point prompts positioned randomly anywhere on the ground truth mask.
    \item \textbf{Bounding Box with positive and negative point prompts (BBOXP1-1, BBOXP4-4)}: To assist SAM in excluding ``holes'' within the iris and sclera segmentations, we provide it with a tight bounding box and a single or 4 positive point prompts placed randomly on the ground truth mask (the same as BBOXP1 and BBOXP4). We furthermore provide SAM with the same number of negative point prompts placed on the ground truth iris mask (when segmenting the sclera) or on the pupil mask (when segmenting the iris). In this experiment, images where the ``holes'' are not present (i.e. in cases where the eye is half-covered by the eyelid or in the process of blinking such that the sclera is visible but not the iris, or the iris is visible but not the pupil) are skipped. This strategy furthermore does not apply to pupil segmentation. 
    
\end{enumerate}

Furthermore, we explore the impact on SAM's performance when introducing variability in the bounding box strategy (BBOX). Taking inspiration from \citeauthor{mattjie2023zeroshot}~[\citeyear{mattjie2023zeroshot}], we adjust the bounding box around the ground truth mask, modifying its size and position by 5\%, 10\%, and 20\%. This variation mirrors the potential differences that may arise when different annotators draw bounding boxes over the image.

Across all strategies, images where the targeted feature is absent are excluded. In other words, if SAM's ability to segment a specific feature (e.g., pupil) is being evaluated, images in which that feature is not present (e.g., a half-lidded eye that only shows the sclera and iris but not the pupil) are skipped.

\subsection{Evaluation}
To evaluate SAM's segmentation quality, we want to analyze the overlap between its segmentations and the ground truth masks, while also gauging the alignment of the two masks. To measure this, we chose well-established metrics that are commonly used in evaluating image segmentation models:

\textbf{Dice similarity coefficient (Dice)} \cite{milletari2016v}: a similarity metric that gauges the overlap between the ground truth mask and the predicted mask. A higher value signifies better model performance. Given two sets of finite points A and B, the Dice coefficient is defined as: 
\begin{equation}
    Dice = \frac{2 \times |A \cap B|}{|A| + |B|}
\end{equation}

\textbf{Intersection-over-union (IoU) }\cite{zhou2019iou}: similar to Dice, it measures the overlap between two masks by getting the ratio of the intersection area to the union area of the segments. IoU requires more precision than Dice as it penalizes both under- and over-segmentation more rigorously. Higher values signify a better model:
    \begin{equation}
    \text{IoU} = \frac{{|A \cap B|}}{{|A \cup B|}}
\end{equation}

 \textbf{Hausdorff Distance (HD)} \cite{huttenlocher1993comparing}: a boundary-based metric that quantifies the dissimilarity between two shapes by measuring the maximum distance between points on one boundary to the nearest points on the other boundary. This metric can be used to evaluate the degree of alignment between the boundaries of the predicted and ground truth masks. A lower Hausdorff distance indicates better model performance. Given two sets of finite points A and B and with $d(a,b)$ as the distance between points a and b, the Hausdorff Distance is defined as: 
    \begin{equation}
        H(A, B) = max(h(A, B), h(B, A)) 
    \end{equation}
    where 
    \begin{equation}
        h(A, B) = \sup_{a \in A} \inf_{b \in B} \, d(a, b)
    \end{equation}
where $\sup$ represents the supremum operator and $\inf$ the infimum operator.

\section{Results}

\begin{figure}[htbp]
    \centering
    \begin{minipage}{0.79\textwidth} 
        \centering
        \includegraphics[width=\textwidth]{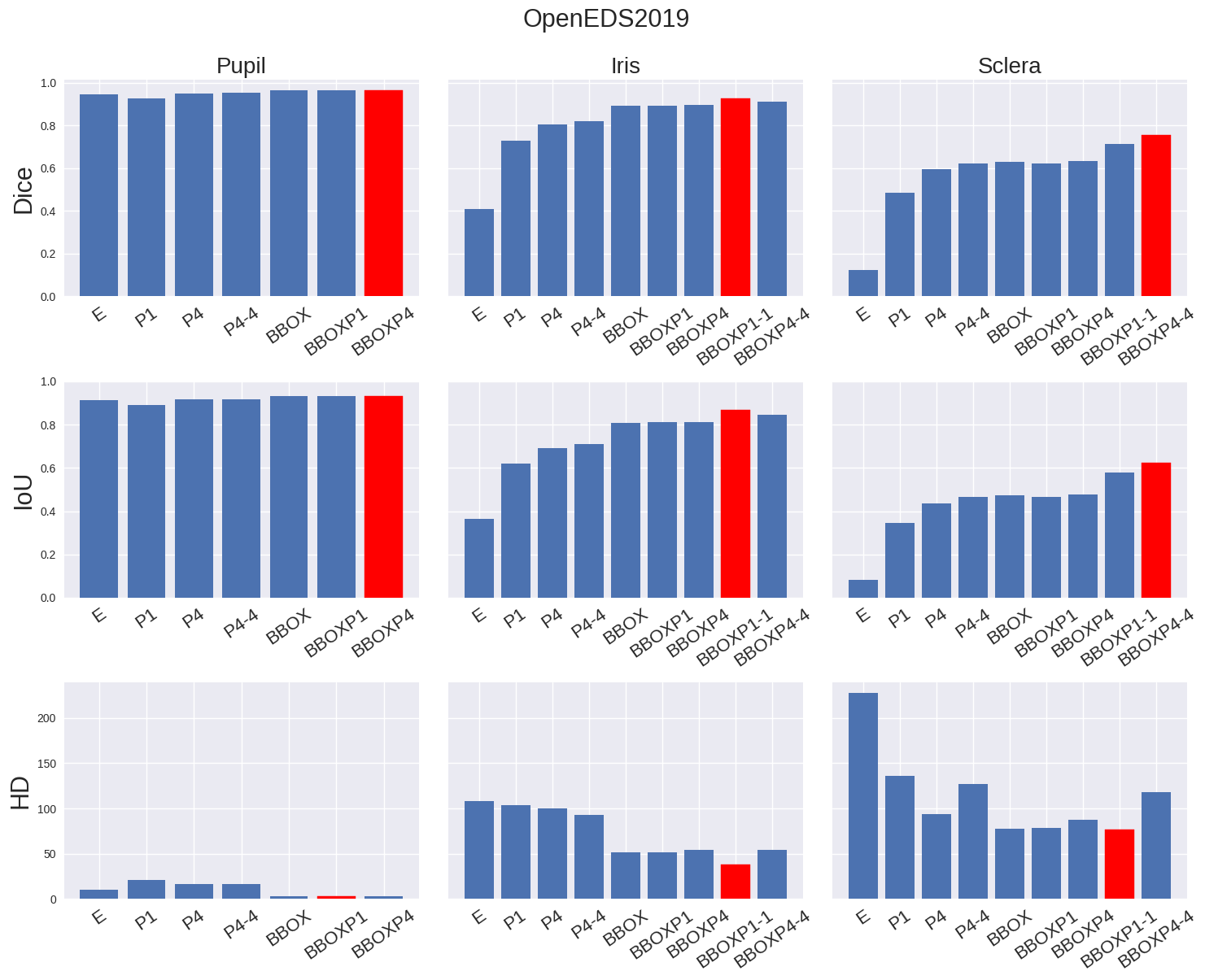} 
        \label{fig:eds19results}
    \end{minipage}
    \begin{minipage}{0.79\textwidth} 
        \centering
        \includegraphics[width=\textwidth]{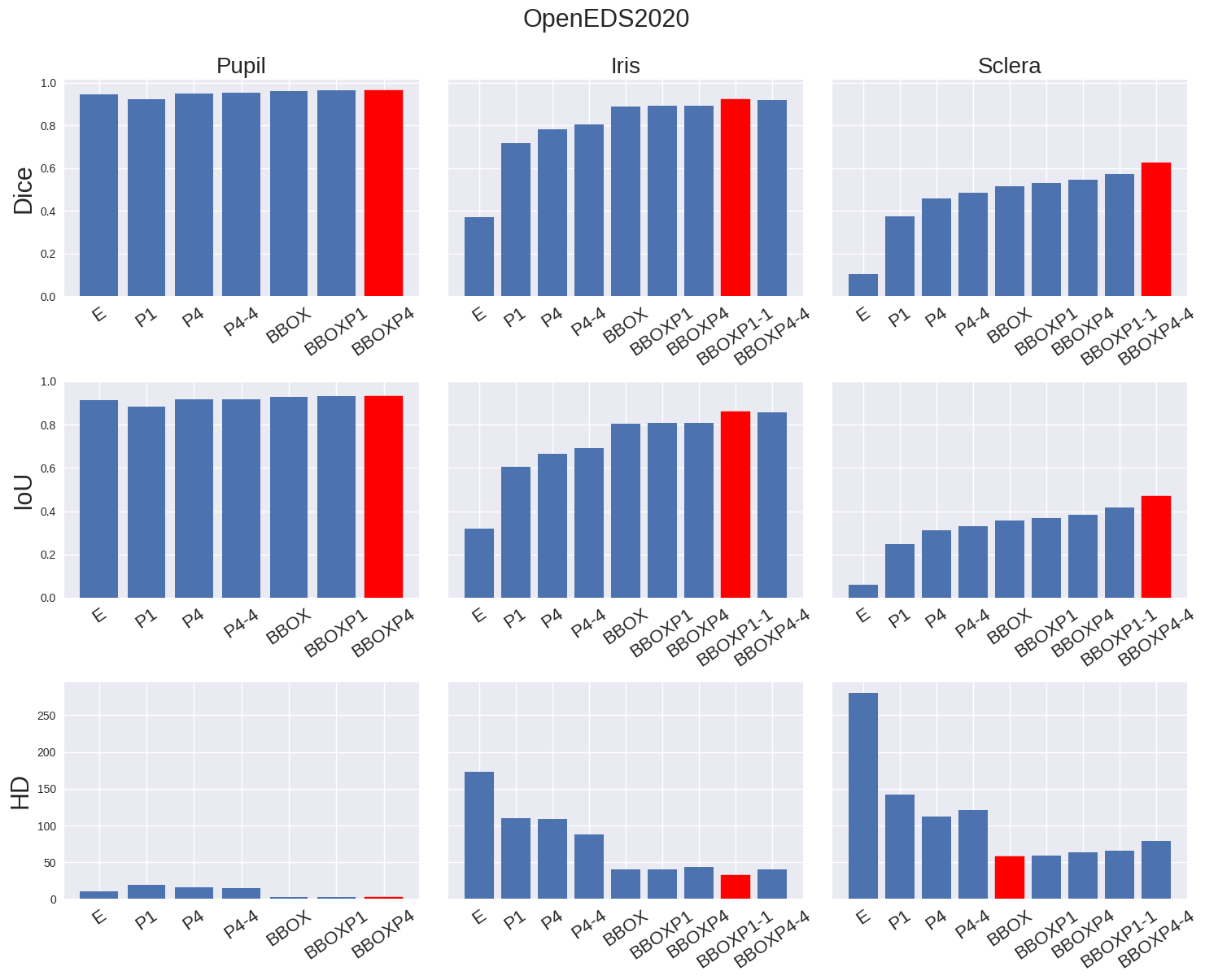} 
        \label{fig:eds20results}
    \end{minipage}
    \caption{Performance of each strategy on different segmentation tasks (pupil, iris, and sclera segmentation). Each row represents a different metric. On each plot, the best-performing strategy (highest Dice/IoU and lowest HD) is colored red.}
    \label{fig:both}
\end{figure}

\begin{table}[]
\begin{tabular}{@{}lcccllllll}

 \multicolumn{10}{c}{\textbf{OpenEDS-2019}}\\
 \toprule
 & \multicolumn{3}{c}{Pupil}& \multicolumn{3}{c}{Iris}& \multicolumn{3}{c}{Sclera}\\
\toprule 

 Perturbance& 5\%& 10\%& 20\%& 5\%& 10\%& 20\%& 5\%& 10\%&20\%\\
 \midrule 
 \textbf{Dice $\uparrow$}  & 0.9546& 0.8423& 0.4493
& 0.8916& 0.8775& 0.7165
& 0.6184& 0.6021&0.4574
\\
 \textbf{IoU $\uparrow$}   & 0.9194& 0.7903& 0.4086
& 0.8083& 0.7917& 0.6261
& 0.4616& 0.4493&0.3330
\\
 \textbf{HD $\downarrow$}                            & 4.1031& 12.1731& 37.5265& 52.4545& 52.9094& 63.3940& 78.6238& 79.8332&92.8560\\
\toprule 
\\
 \multicolumn{10}{c}{\textbf{OpenEDS-2020}}\\
\toprule
& \multicolumn{3}{c}{Pupil}& \multicolumn{3}{c}{Iris}& \multicolumn{3}{c}{Sclera}\\
\toprule
         Perturbance& 5\%& 10\%& 20\%& 5\%& 10\%& 20\%& 5\%& 10\%&20\%\\ \midrule
\textbf{Dice $\uparrow$}  & 0.9458&  0.7849&0.3583
& 0.8887& 0.8763& 0.7795
& 0.5124& 0.4701&0.2932
\\
\textbf{IoU $\uparrow$}   & 0.9029&  0.7229&0.3244
& 0.8038& 0.7895& 0.6830
& 0.3526& 0.3213&0.1967
\\
\textbf{HD $\downarrow$}                            & 4.8235&  13.9860&37.8557& 40.7677& 42.4274& 54.2700& 58.6688& 72.2898&144.7311\\
\end{tabular}
\caption{SAM's performance on OpenEDS-2019 and -2020 datasets when perturbing the bounding box by 5\%, 10\%, and 20\%}
\label{tab:perturbedBbox}
\end{table}

Our findings across both datasets demonstrate that SAM's performance is highly dependent on the type of feature it is segmenting. When segmenting the pupil (refer to Fig. \ref{fig:both}, left column), SAM consistently had an IoU score above 88\% across both datasets in all strategies, with BBOXP4 yielding the best results with an IoU score of 93.30 \% and 93.34\% for OpenEDS-2019 and OpenEDS-2020 respectively. When segmenting the iris (Fig. \ref{fig:both}, middle column), E performed far more poorly with an IoU score of 36.45\% for OpenEDS-2019 and 31.92\% for OpenEDS-2020. Meanwhile, the prompting strategies all scored above 60\% and BBOXP1-1 was the best-performing strategy with an IoU score of 86.63\% and 86.14\% for OpenEDS-2019 and OpenEDS-2020 respectively. Lastly, when segmenting the sclera (Fig. \ref{fig:both}, right column), SAM performed very poorly compared to when segmenting the pupil and sclera with an IoU score of 8.39\% and 5.86\% when using the automatic mode. While introducing manual prompts boosted its performance, it reached only an IoU score of 62.19\% and 47.06\% using BBOXP4-4 for OpenEDS-2019 and OpenEDS-2020 respectively, indicating the need for additional manual guidance. 

\begin{figure}[h]
    \centering
    \begin{minipage}[t]{\textwidth}
        \centering
        \includegraphics[width=\linewidth]{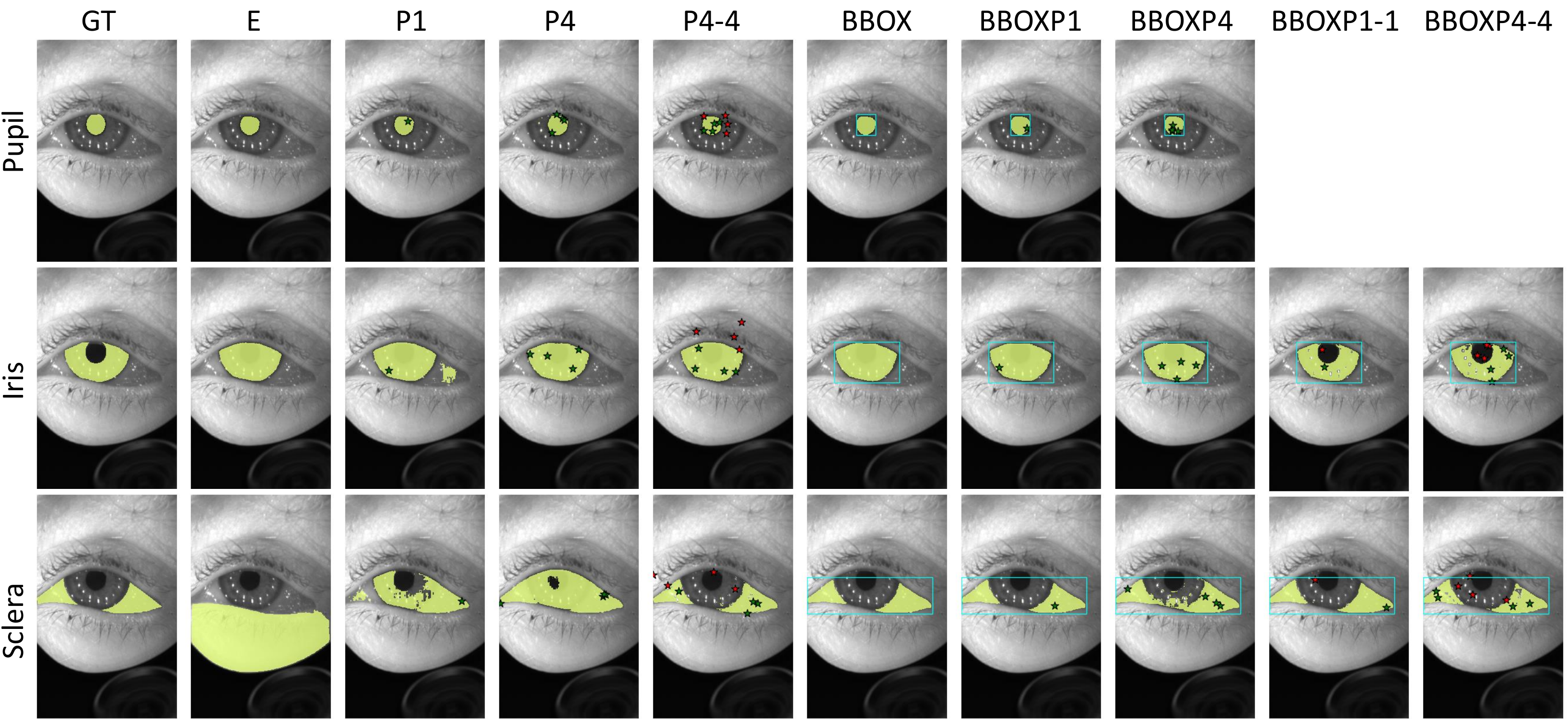}
        \caption{Visualization of SAM's performance on an image from OpenEDS2019 \cite{garbin2020dataset}. The leftmost column shows the ground truth masks for pupil (top row), iris (middle row), and sclera (bottom row) overlayed on the input image. The remaining columns show SAM's segmentations using different strategies.}
        \label{fig:eds19_qualitative}
    \end{minipage}
    
    \vspace{0.1cm} 
    \begin{minipage}[t]{\textwidth}
        \centering
        \includegraphics[width=0.7\linewidth]{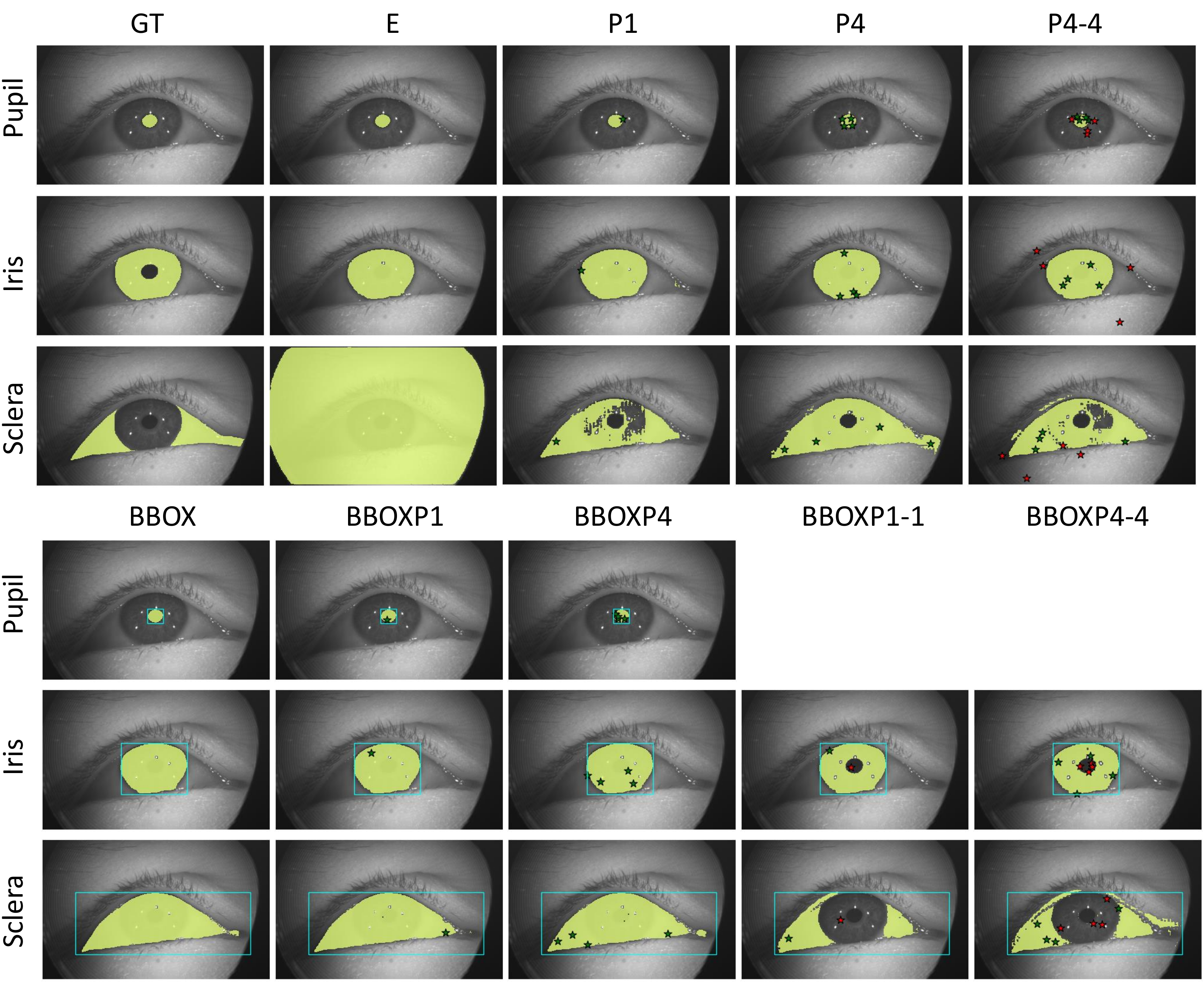}
        \caption{Visualization of SAM's performance on an image from OpenEDS2020 \cite{palmero2021openeds2020}}
        \label{fig:eds20_qualitative}
    \end{minipage}
\end{figure}

Comparing across prompting strategies (Fig. \ref{fig:both}), E performed reasonably well when segmenting the pupil and even outperformed P1 in both datasets. However, its performance decreased dramatically when segmenting the iris and sclera, in some cases, not overlapping at all with the ground truth mask. This is exemplified by E's predictions on the sclera in Fig. \ref{fig:eds19_qualitative}, and \ref{fig:eds20_qualitative} which show in both cases that the closest matching segmentation that SAM produces is nowhere near the ground truth mask of the sclera. When introducing point prompts (P1, P4, P4-4), SAM's performance increased with the number of point prompts used. Using a simple bounding box prompt (BBOX), however, outperformed the use of only point prompts. This is likely because the bounding box more explicitly delineates the area where the feature of interest is located, while point prompts may be ambiguous regarding exactly which object the prompt is referring to. Lastly, a combination of bounding box and point prompts (BBOXP1, BBOXP4, BBOXP1-1, and BBOXP4-4) resulted in the best performance of SAM on both datasets for both the iris and sclera.  

Analyzing the impact of perturbed bounding boxes (refer to Table \ref{tab:perturbedBbox}), SAM demonstrates a decline in performance with decreasing accuracy of the bounding box placement. However, even when the bounding box is perturbed, SAM still consistently outperforms the majority of the point prompt strategies when segmenting the iris and sclera, particularly with perturbations of 5\% and 10\%. This observation aligns with the findings of \citeauthor{mattjie2023zeroshot}~[\citeyear{mattjie2023zeroshot}], underlining the robustness of the bounding box approach compared to point prompts. 

When evaluating SAM's performance based on the alignment of mask boundaries (refer to Fig. \ref{fig:both}, third and last rows), it is evident that strategies incorporating the bounding box prompt consistently yield the best results. However, it is noteworthy that not all strategies deliver the highest degree of alignment (measured by the lowest HD) and the best overlap (measured by the highest IoU and Dice scores) simultaneously. This is particularly evident when examining performance on the sclera. While BBOXP4-4 showed the best overlap for both datasets, it did not show the best alignment. Meanwhile, the BBOXP1-1 strategy showed the best alignment for OpenEDS2019 and BBOX performed best on OpenEDS2020 while these showed up to almost 30\% worse overlap scores.  This indicates that, although incorporating point prompts with the bounding box prompt may help SAM to capture the general shape of the sclera, it might actually hurt its ability to accurately delineate its exact boundaries. Although less clear, the same trend is seen in the iris data, where adding a single prompt point to the bounding box lead to better alignment than adding four prompt points.

Choosing the best-performing strategy for the pupil (BBOXP4), iris (BBOXP1-1), and sclera (BBOXP4-4), we calculated the mIoU for each dataset. The results indicated a mIoU of 85.70\% and 83.90\% on the OpenEDS-2019 and OpenEDS-2020 datasets, respectively. These values fell short of the previously established baselines of 91.4\% ~\cite{garbin2020dataset} and 84.1\% ~\cite{palmero2021openeds2020}. This is likely attributed to SAM's limitations in accurately segmenting the sclera as previously observed. Indeed, upon excluding the results related to sclera, the mIoU improved significantly to 91.38\% for OpenEDS2019 and 91.02\% for OpenEDS2020. However, these scores still lag behind the top-performing results from the accompanying leaderboards\footnote{https://eval.ai/web/challenges/challenge-page/353/leaderboard/1002}\textsuperscript{,}\footnote{https://eval.ai/web/challenges/challenge-page/603/leaderboard/1680}, which achieved a mIoU score of 95.28\% for OpenEDS2019 and 95.17\% for OpenEDS2020. Including this comparative analysis, our results underscore the need for further refinement of the model in order to match with the top-performing models. However, it is noteworthy that the SAM model has not been previously exposed to near-eye image datasets, in contrast to the compared models that were trained on the OpenEDS datasets themselves, hinting at its potential generalizability to various eye tracking datasets.

\section{Discussion}

Evaluating the capability of the Segment Anything Model (SAM) in delineating critical eye features for eye tracking, our study reveals that SAM effectively performs zero-shot segmentation on the pupil, achieving an impressive IoU (Intersection over Union) of 93.34\%. With the assistance of bounding box and point prompts, it also successfully segments the iris, reaching an IoU of 86.63\%. Nevertheless, SAM encounters difficulties with the sclera, achieving a best IoU of only 62.19\%, and this limitation persists regardless of the type of prompt used, even though prompts do enhance the performance over its automatic mode. These findings are consistent with previous investigations mainly from the medical imaging sector~\cite{mazurowski2023segment,huang2023segment}. Furthermore, it is notable that SAM's performance in iris segmentation does not significantly improve with multiple point prompts when combined with a box prompt, as opposed to adding a single prompt. This observation corresponds with previous literature \cite{mazurowski2023segment}, indicating that additional point prompts do not generally enhance SAM's performance, except in cases where the target has multiple parts, such as the sclera.

The fundamental question for those in the eye-tracking field may boil down to ``Is SAM suitable for segmenting eye images?'' The answer hinges on the specific feature in question. For example, SAM exhibits strong capabilities in pupil segmentation which can be used in pupil detection pipelines, a crucial element in numerous gaze-tracking frameworks~\cite{kim2019nvgaze}, and the primary subject of various studies~\cite{fuhl2016pupilnet,santini2018pure,fuhl2015excuse}. The effectiveness of SAM for segmenting the iris is comparatively lower, compared to the pupil. We attribute this reduction in performance to the iris having a lower contrast and relatively blurred edge with the sclera, conditions that have previously been observed to pose problems for SAM~\cite{mattjie2023zeroshot,huang2023segment}. Moreover, our results reveal that SAM struggles to recognize the sclera as a distinct object and requires extensive guidance from an annotator through the use of manual prompts. We speculate that shadows from the eyelids and non-uniform illumination contribute to large variations in scleral luminance in the images. This, combined with the blurry edge between the iris and the sclera, likely poses a significant challenge for SAM. 

\section{Limitations and Future Work}

Our study presents several limitations. First, our study assessed only the zero-shot capabilities of SAM within a Virtual Reality (VR) environment. Future research should extend the evaluation of SAM's performance to eye images acquired from wearable as well as high-resolution laboratory eye trackers. High-resolution eye tracking may demand greater precision and the segmentation of different features~\cite{byrne2023leyes} for which the segmentation techniques utilized in our study may not be effective. The complexity of ``gaze-in-the-wild'' contexts, characterized by considerable variability in eye appearance and challenges with low image quality, may present a more arduous challenge than VR environments investigated in this paper~\cite{kothari2020gaze,fuhl2016pupil,fuhl2022pistol}. Second, there is potential for uncovering more effective prompting strategies and image augmentation techniques that could further enhance the model's performance in segmenting eye regions. Finally, our study did not address the segmentation of corneal reflections, despite their presence in the images (as demonstrated in \cite{chugh2021detection,maquiling2023v,byrne2023leyes}) because they were not annotated in the dataset. Future work should consider including these reflections, as the detection and matching of these features can be integral to comprehensive eye-tracking analyses and P-CR pipelines~\cite{chugh2021detection}. 

This research opens the door to a variety of future research avenues. One promising direction is the fine-tuning of SAM on a small amount of eye images, to investigate if this fine-tuning step improves the results compared to the standard model. Second, developing a foundation model trained on a comprehensive dataset, such as TEyeD~\cite{Fuhl_2021}, could become a pivotal asset for the eye-tracking community. However, this endeavor would necessitate computational resources beyond those typically accessible to a standard laboratory. Another intriguing potential research path involves integrating text prompting into the model to further simplify the annotation process for users without technical expertise.

\section{Conclusion} 
 This study explored the applicability of the Segment Anything Model (SAM) for segmenting eye images, the first study into a foundation model's performance in this specific task to our knowledge. SAM showed promising zero-shot capabilities, particularly in pupil segmentation. While we recommend a cautious approach to using SAM for eye image segmentation in its current state, this technology may represent a paradigm shift away from specialized supervised learning models towards foundational models. However, what the eye-tracking community may need for this technology to be fully realized is a bespoke foundation model trained on eye images, akin to how MedSAM~\cite{ma2024segment} is tailored for medical images or RETFound~\cite{zhou2023foundation} for retinal imaging. Regardless, foundation models offer an exciting new direction for the field. They could reduce the technical barriers associated with developing specialized models, mitigate the acute shortage of annotated datasets required for training deep learning systems, and achieve domain generalization—critical challenges in gaze estimation. This shift has the potential to democratize eye-tracking technologies, particularly as VR systems become increasingly mainstream, enabling new market entrants to compete with established companies by leveraging the scalable utility of foundation models.
\bibliographystyle{ACM-Reference-Format}
\bibliography{sample-base}

\end{document}